\documentclass{article}

\usepackage{arxiv}
\usepackage[utf8]{inputenc} 
\usepackage[T1]{fontenc}    
\usepackage{hyperref}
\usepackage{url}            
\usepackage{booktabs}       
\usepackage{amsfonts}       
\usepackage{nicefrac}       
\usepackage{microtype}

\usepackage{graphicx}
\usepackage{natbib}
\usepackage{subfig}
\usepackage{stfloats}

\usepackage{amsmath}
\usepackage{amssymb}
\usepackage{mathtools}
\usepackage{amsthm}
\usepackage{multirow}
\usepackage{multicol}
\usepackage{makecell}
\usepackage{enumitem}

\usepackage{cleveref}
\theoremstyle{plain}

\theoremstyle{definition}

\theoremstyle{remark}

\usepackage[textsize=tiny]{todonotes}

\title{Revisiting the robustness of post-hoc interpretability methods}

\date{}

\newif\ifuniqueAffiliation
\uniqueAffiliationtrue

\ifuniqueAffiliation 
\author{{Jiawen Wei} \\
	Department of Mechanical Engineering\\ 
        College of Design and Engineering\\
        National University of Singapore, Singapore\\
	\texttt{jiawenw@u.nus.edu} \\
	\And
    {Hugues Turbé} \\
	Division of medical information sciences\\
	University hospitals of Geneva and \\
        Department of radiology and medical informatics\\
        University of Geneva, Switzerland\\
	\texttt{hugues.turbe@unige.ch} \\
 	\And
 {Gianmarco Mengaldo}\thanks{Corresponding author: mpegim@nus.edu.sg} \\
	Department of Mechanical Engineering\\ 
        College of Design and Engineering\\
        National University of Singapore, Singapore\\
	\texttt{mpegim@nus.edu.sg} \\
}
\else
\usepackage{authblk}

\newbox{\orcid}\sbox{\orcid}{\includegraphics[scale=0.06]{orcid.pdf}} 
\author[1]{%
	\href{https://orcid.org/0000-0000-0000-0000}{\usebox{\orcid}\hspace{1mm}David S.~Hippocampus\thanks{\texttt{hippo@cs.cranberry-lemon.edu}}}%
}
\author[1,2]{%
	\href{https://orcid.org/0000-0000-0000-0000}{\usebox{\orcid}\hspace{1mm}Elias D.~Striatum\thanks{\texttt{stariate@ee.mount-sheikh.edu}}}%
}
\affil[1]{Department of Computer Science, Cranberry-Lemon University, Pittsburgh, PA 15213}
\affil[2]{Department of Electrical Engineering, Mount-Sheikh University, Santa Narimana, Levand}
\fi

\renewcommand{\shorttitle}{Revisiting the robustness of post-hoc interpretability methods}

\hypersetup{
pdftitle={Revisiting the robustness of post-hoc interpretability methods},
pdfsubject={q-bio.NC, q-bio.QM},
pdfauthor={Jiawen Wei, Hugues Turbé, and Gianmarco Mengaldo},
pdfkeywords={First keyword, Second keyword, More},
}

\begin{document}

\maketitle

\begin{abstract}
Post-hoc interpretability methods play a critical role in explainable artificial intelligence (XAI), as they pinpoint portions of data that a trained deep learning model deemed important to make a decision. 
However, different post-hoc interpretability methods often provide different results, casting doubts on their accuracy. 
For this reason, several evaluation strategies have been proposed to understand the accuracy of post-hoc interpretability. 
Many of these evaluation strategies provide a coarse-grained assessment -- i.e., they evaluate how the performance of the model degrades on average by corrupting different data points across multiple samples. While these strategies are effective in selecting the post-hoc interpretability method that is most reliable on average, they fail to provide a sample-level, also referred to as fine-grained, assessment. In other words, they do not measure the robustness of post-hoc interpretability methods.
We propose an approach and two new metrics to provide a fine-grained assessment of post-hoc interpretability methods. We show that the robustness is generally linked to its coarse-grained performance.
\end{abstract}

\section{Introduction} 
\label{sec:intro}
The increasingly widespread adoption of deep learning models across different fields is pushing an increasing demand for artificial intelligence (AI) transparency~\cite{ai_act}. 
Transparency can be defined at many levels of the AI workflow, ranging from AI awareness, to AI model interpretability, and AI outcome interpretability. 
The focus of this work is on the latter -- AI outcome interpretability. This aspect of transparency is concerned with explaining the input contributions to a given output. 
To achieve AI outcome transparency, several approaches have been proposed. These are usually categorized in ante-hoc (also referred to as self-explainable) and post-hoc interpretability. The former try to build AI models that are interpretable by construction -- e.g.,~\cite{li2018deep,turbe2024protos}. 
The latter refer to methods that allow probing black-box AI models and grasp an approximate view of what inputs where used for a certain output -- e.g.,~\cite{ribeiro2016model}. 
Both ante-hoc and post-hoc interpretability methods can be used to assess whether decisions made by an AI model are trustworthy -- i.e., the converge to human domain experts understanding of a given problem~\cite{mengaldo2024explain}. 
This aspect may be useful for regulatory purposes in critical sectors such as the healthcare. 
They may also be used to shape AI model behaviour~\cite{nickl2024memory}, by e.g., completing the dataset with relevant adversarial samples.  
More recently, it was additionally proposed that interpretability methods can also be useful for knowledge discovery, especially when the AI model provides a divergent view from human domain experts, where the latter can try to explain why the machine used certain inputs to reach a certain outcome~\cite{mengaldo2024explain}.

In this work, we focus on post-hoc interpretability methods. These methods were initially proposed for image classification tasks~\cite{zeiler2014visualizing}, and they were later applied to text and time-series classification tasks~\cite{samek2017explainable,ismail2020benchmarking}.
In recent years, this field of interpretability has gained significant traction, leading to several model-agnostic methods to understand the contribution of input features to a given AI output~\cite{zhang2021survey}. An example of such traction is the open-source library Captum~\cite{kokhlikyan2020captum}, an effort brought forward by Meta to encapsulate post-hoc interpretability methods under a unique software framework.

However, despite the significant interest in this field, a critical issue remains: the discrepancy that different post-hoc interpretability methods provide when interpreting the same trained model on the same sample of a given dataset. 
For such task, all the different interpretability methods available should provide the same result, yet they frequently provide  different relevance maps (maps of the input data deemed important by the AI model).
This critical issue should be solved or at least mitigated, before moving towards reliable and useful post-hoc interpretations of deep learning models~\cite{lipton_mythos_2017,rudin2022interpretable}.

To mitigate and possibly address this issue, a few evaluation strategies of interpretability methods have been proposed in the literature. These are commonly based on the notion of corrupting (also referred to as occluding in computer vision) data points deemed relevant by the interpretability method, and evaluate how the output of the model changes~\cite{turbe2023evaluation}. This allows understanding how accurately a post-hoc interpretability method captures the input data that was actually used by the deep learning model. Hence, they provide a way to rank post-hoc interpretability methods by the accuracy of their explanations.

\begin{figure}[t]
\begin{center}
\centerline{\includegraphics[width=0.65\textwidth]{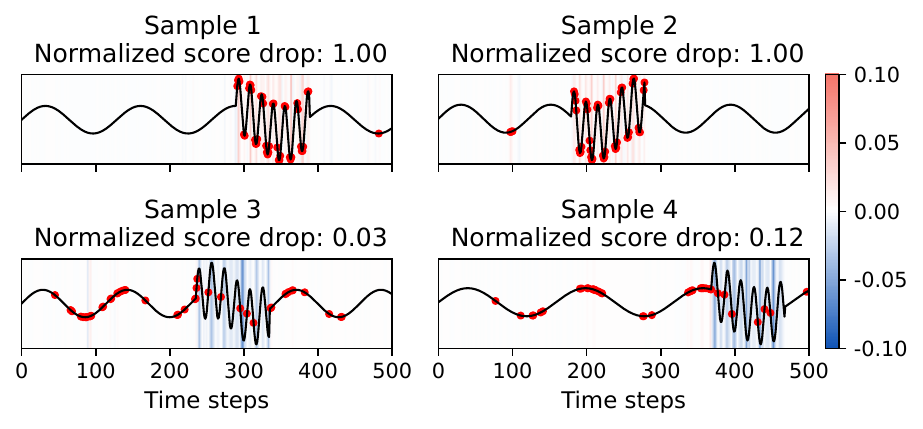}}
\caption{With the \textbf{mean} score drop of 0.66 at top-15\% corruption, DeepSHAP is not robust for all samples in the synthetic dataset.}
\label{fig:intro}
\end{center}
\end{figure}

Yet, these evaluation methods usually take into account average performance across several samples within the dataset of interest. 
This \textit{coarse-grained} evaluation is helpful to assess whether an interpretability method is providing explanations that are accurate on average. 
However, it does not provide a \textit{fine-grained} view of how an interpretability method performs at sample level, and the sensitivity of the explanations for different samples. In other words, they do not assess the \textit{robustness} of post-hoc interpretability methods. 
One may ask why this is important. 
Consider Figure \ref{fig:intro}. 
Here, we show the relevance map (i.e., the data portions deemed most important for the output of the trained model) produced by a post-hoc interpretability method, namely DeepSHAP~\cite{lundberg2017unified}, for a time series classification task. 
The red dots pinpoint the top-15\% most important data portions for four different samples. 
On top of each subfigure, we report the \textit{normalized score drop} -- i.e., how much the output changed after corrupting the red dots. 
We note that for sample 1 and 2, the score drop is significant, while for sample 3 and 4, is extremely small. 
We also note a different clustering of red dots in the top two plots vs the bottom two plots. In the top two plots, the red dots are clustered in the high-frequency wave-pocket that constitutes the discriminative part for the learning task at hand. 
In the bottom two plots the red dots are distributed across the entire time series. 
If we were to compute an average score drop for DeepSHAP, we would obtain a value of 0.66. 
Yet, the performance of DeepSHAP for the top two vs the bottom two plots is remarkably different. 
Indeed, for sample 3 and 4, DeepSHAP seem to fail capturing the relevant portions of the data used by the deep learning model to classify the time series. 

In this work, we propose a new methodology for deep learning classification that provides a \textit{fine-grained} understanding of post-hoc interpretability methods performance, thereby assessing their \textit{robustness}. 
We frame \textit{robustness} as the ability of post-hoc interpretability methods to pinpoint the relevant portion of data across all (or at least the majority) of the samples in a given dataset. 
This means having score drops that do not vary widely across samples, like in Figure \ref{fig:intro}. 
We note that the score drop depends on two factors: (i) the ability of the interpretability method to capture the most relevant portion of the data, and (ii) the sensitivity of the sample in question to data corruption or in other words, to how close the sample is to the decision boundary of the classification task. 
The first point provide a fine-grained quantification of how different post-hoc interpretability methods perform across samples, and it is the main target of this study. The second point might be used to grasp the sensitivity of each sample to corruption of the input data. To this end, we performed a calibration study and some tests in the appendix, that sheds light into this aspect.
%
We show that we can use ridge-line plots (see Figure \ref{fig:ridge_skew_kurt}) as a fine-grained tool to visualize the distribution of the score drop across samples. 
The properties of the distribution provide quantitative metrics for assessing the robustness of different interpretability methods. 
In the literature, one commonly takes the average of the distribution to provide a coarse-grained assessment of interpretability methods. 
Indeed, we use two metrics, $\operatorname{AUC}\tilde{S}_\text{top}$ and $\operatorname{F1}\tilde{S}$, first proposed in~\cite{turbe2023evaluation} to measure the average performance of interpretability methods -- i.e., to perform a coarse-grained assessment. 
These two metrics use the area under the curve (AUC) and the harmonic mean (F1) of normalized score drops for different number of corrupted input data points. 
Here, we propose two complementary metrics, namely $\overline{\operatorname{AUC}\textit{Skew}}$, to quantify the \textit{robustness} of post-hoc interpretability methods. 
These two metrics are based on higher order moments of the distribution, i.e., \textbf{skewness} and \textbf{kurtosis}, respectively. 
The evaluation workflow, depicted in Figure \ref{fig:framework}, should first include a coarse-grained assessment, to understand average performance, followed by a fine-grained assessment if the average performance are satisfactory. 
It is hardly useful to perform a fine-grained assessment using the two new metrics we propose if the average performance are poor. 

The methodology that we propose is general and can be applied to any type of data, including images, time series and text. In this paper, we show results for time series classification tasks~\cite{dhariyal2023back}, as they represent a relatively challenging application for interpretability~\cite{turbe2023evaluation,ismail2019deep,assaf2019mtex,montavon2018methods,di2023explainable}, that is only partially explored.


Our key contributions are summarized as follows:
\begin{itemize}[leftmargin=*, topsep=0pt, partopsep=0pt, itemsep=0pt]
    \item We define \textit{robustness} as the ability of post-hoc interpretability methods to identify the data portions used by the model to make its predictions consistently across samples. This means looking at the properties of the distribution of different samples in terms of \textit{predicted probability changes} after corrupting the portions of data deemed most important by the interpretability method. 
    \item We propose \textbf{ridge-line plots} as a fine-grained tool to visualize the distribution of \textit{predicted probability changes}.
    \item We propose \textbf{two new fine-grained metrics}, $\overline{\operatorname{AUC}\textit{Skew}}$ and $\operatorname{AUC}\textit{Kurt}$, to quantitatively evaluate \textit{robustness}.
    \item We argue that a framework from coarse- to fine-grained evaluation is necessary for the correct assessment of post-hoc interpretability methods. 
    \item We conduct experiments on one synthetic and 20 public datasets to validate our evaluation framework. 
    \item We show how the sensitivity of the samples  displays an important interplay with interpretability methods and their evaluation; aspect that should be taken into account.
\end{itemize}

The rest of this paper is organized as follows. Section~\ref{sec:2} briefly summarizes recent competing work related to post-hoc interpretability evaluation and robustness. Section~\ref{sec:3} introduces our evaluation framework in detail, including ridge-line visualization and new fine-grained quantitative evaluation metrics. Section~\ref{sec:4} thoroughly presents our experimental findings and exploratory analysis. Section~\ref{sec:5} concludes with some remarks and discusses potential future work.

\section{Relevance and related work} 
\label{sec:2}

\subsection{Relevance}
\label{subsec:relevance}
In this work, we rethink post-hoc interpretability robustness. In particular, we look at how robust post-hoc explanations are across samples, probing the properties of the score drop distribution arising of all samples in the dataset of interest. While the idea seems relatively trivial, this aspect has not been considered in the literature. Yet, it is extremely important to understand the robustness properties of interpretability methods, especially if we are interested in single samples of our data. Indeed, this is often the case when demanding interpretability results. 

Consider the case of a patient, who might be affected by a certain heart disease. 
If we use electrocardiogram (ECG) data to assess the potential disease of the patient, we can setup a classification task that probes the ECG time series and classifies them as belonging to a disease type. 
The subsequent task is to pass the results to a clinician (human-in-the-loop) who needs to make a diagnosis based on the AI results and the ECG data. 
At this point, the clinician may ask: what data the AI model used to take its decision \textit{for this particular patient}? Different post-hoc interpretability methods will provide different answers to this question. 
Hence evaluating the accuracy of the explanations is as important as interpretability per se. 
However, this evaluation is commonly done considering the \textit{average} accuracy of post-hoc interpretability on the entire patient population. 
This approach discounts other properties of the distribution, in addition to the average. 
We argue that these other properties are equally important, especially if we want to evaluate how consistent the interpretability results are across different samples (e.g., patients). 
In other words, they provide a measure of uncertainty regarding the explanations provided. 
Going back to the heart disease example, this means that we are now able to choose the interpretability method that gives the best results on average but that also provides robustly accurate (or less uncertain) explanations across different patients. 
This is of course critically important if we want to provide an accurate answer to the original question asked by the clinician. 
More specifically, it allows the clinician to view the interpretability results with associated uncertainty, aspect that is of paramount importance in  critical sectors. 

Indeed, in high-risk application sectors, such as medicine, and finance, non-robust explanations may limit the use of deep learning solutions. 
This because non-robust explanations may undermine the pillars of deep learning deployment commonly accepted by humans, including: trust, informativeness, and fair and ethical decision-making~\cite{lipton_mythos_2017}. 
Pillars that are now being increasingly regulated~\cite{ai_act}. 

In addition, having a granular understanding of the correct portions of input data used by deep learning models across samples can help to improve our knowledge of a given problem. 
It can, for instance, equip domain-expert with new ways of looking at a problem, thereby possibly uncovering causal relationships that were previously not understood.

\subsection{Related work}
\label{subsec:RWB}

The importance of evaluating interpretability has recently received increasing attention, leading to various attempts at evaluating post-hoc interpretability methods. 
A common strategy for evaluation is to remove portions of the input data deemed important for the prediction task to see the degradation of model performance \cite{samek2016evaluating} -- i.e., to see the change in probability (also referred to as normalized score drop) associated to the class identified by the model.
If taken as is, without further considerations, this approach violates the i.i.d.\ assumption since the training and testing datasets generally come from different distributions.
Hence, the decrease in model accuracy might not be due to the removal of relevant portions of input data, but because of the distribution shift between the training and the test dataset.

The ROAR approach (RemOve And Retrain) \cite{hooker2019benchmark} was introduced to address this issue. 
This approach retrains the deep learning model and evaluates the drop in score after occluding relevant portions of the input data. 
While ROAR retains a similar distribution between training and testing, it evaluates the retrained model rather than the initial one used for the prediction task. 

To address these limitations of existing evaluation methods, \cite{turbe2023evaluation} recently introduced data augmentation into the training dataset. 
The augmentation accounts for possible distribution shifts between training and testing, thereby solving the distribution shift and retraining issues. 
The same work also proposes two quantitative metrics $\operatorname{AUC}\tilde{S}_\text{top}$ and $\operatorname{F1}\tilde{S}$ to evaluate post-hoc interpretability methods. 

The works mentioned aim to evaluate the average performance of interpretability methods for a given testing dataset.
Yet, as motivated in section~\ref{subsec:relevance}, it is also critical to understand the behaviour of interpretability methods across samples -- i.e., their robustness (or uncertainty).

There are very few works that attempt to understand robustness of interpretability methods at the sample level. 
One such work introduces metrics to quantify the robustness (the term \textit{robustness} is often used interchangeably to refer to \textit{stability/sensitivity} \cite{jyoti2022robustness}) of local explanation methods, and argue that ``\textit{similar inputs should give rise to similar explanations}"~ \cite{alvarez2018robustness, alvarez2018towards}. 
Built on this assumption, a few researchers measure sensitivity  by calculating the difference in explanations with respect to the change in the input (i.e., infinitesimally small noise to the original instances).
For instance, \cite{yeh2019fidelity} and \cite{bhatt2020evaluating} explore max sensitivity to measure the maximum change in explanations with a small perturbation of the input, while \cite{fel2022good} investigate the dissimilarity between two explanations to evaluate the generalizability and consistency. 
However, these works carry the risk that deep learning models have the same behavior on original and perturbed inputs. 
To overcome this limitation, \cite{agarwal2022rethinking} propose new relative stability metrics to evaluate explanations via three aspects, including change in input, model representation, and output of the predictor. 
Another family of studies bridges adversarial robustness and interpretability evaluation by means of adversarial training or adversarial attacks on the input to measure the deviations from the original explanations~\cite{kim2019bridging, dombrowski2019explanations, heo2019fooling, sinha2021perturbing}. 
In this work, we rethink robustness from a different angle. We evaluate higher order moments of the score-drop distributions, proposing new analytics and metrics to probe and understand post-hoc interpretability performance.

\begin{figure*}[t]
\begin{center}
\centerline{\includegraphics[width=\textwidth]{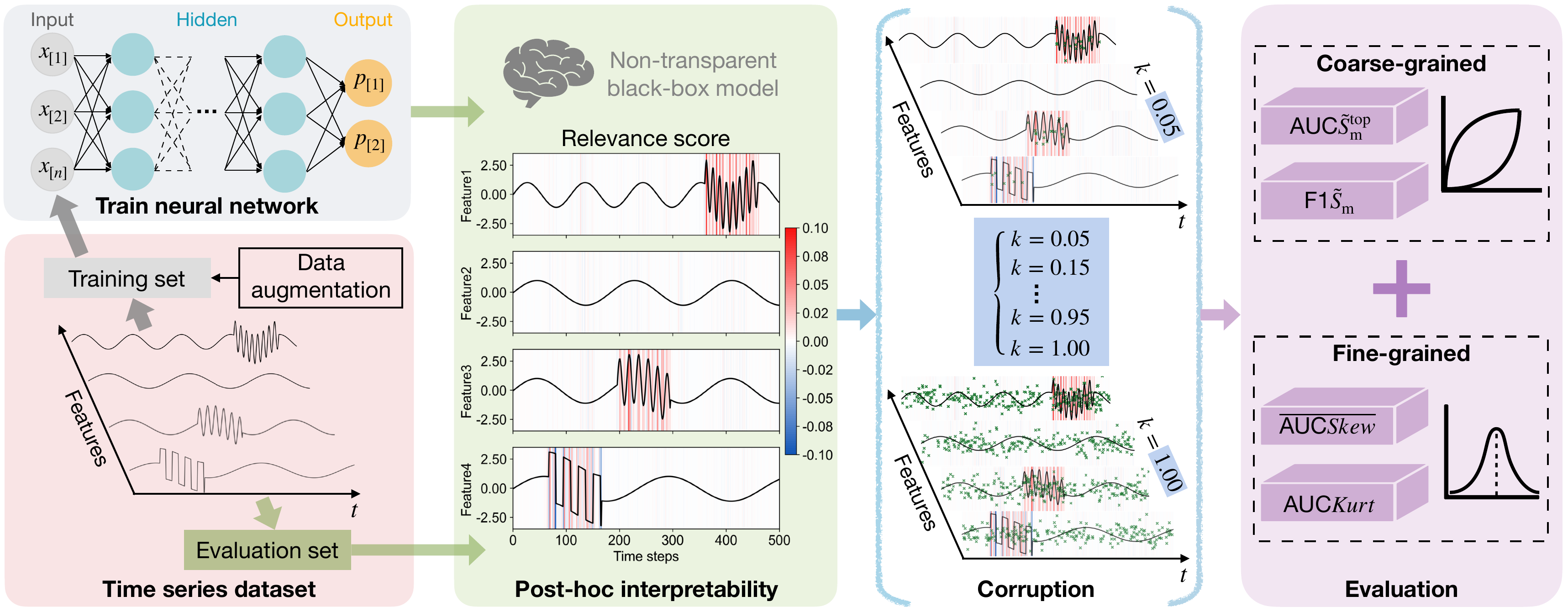}}
\caption{Framework of post-hoc interpretability evaluation consists of four main processes. 1) neural network training for time series classification; 2) post-hoc interpretability methods for relevance score generation in testing set; 3) k-percentile corruption based on relevance ranking; 4) coarse-grained and fine-grained evaluation.}
\label{fig:framework}
\end{center}
\end{figure*}
%

\section{Methodology}
\label{sec:3}

We consider a time series classification dataset $\mathcal{D}=\{\mathrm{X}, c\}$, where $\mathrm{X}$ is the time series input and $c$ is the associated classification label. 
In this setting, each time series is defined as $\mathrm{X}=(x_{m,t}) \in \mathbb{R}^{M \times T}$, where $M$ is the number of features, $T$ is the number of time steps per feature, and $x_{m,t}$ represents the input feature $m$ at time step $t$. 
Taking the time series as the input, neural networks produce a score $\mathcal{S}_c(\mathrm{X})$ for the predicted class $c$. Given class $c$, post-hoc interpretability methods assign relevance scores $\mathrm{R} = (r_{m,t}) \in \mathbb{R}^{M \times T}$ for the input feature $m$ at time step $t$, which can be positive or negative. 
For simplicity, we drop the class $c$ that post-hoc methods aims to interpret for the rest of the paper, and denote the score as $\mathcal{S}(\mathrm{X})$. Those time steps with positive scores (red) are identified to be in favor of the prediction, while the one with negative scores (blue) are against it (see contour map in Figure \ref{fig:framework}).
Here, we focus on positive scores $\mathrm{R}^+:=\{r_{m,t} | r_{m,t} > 0\}$ as we are interested in relevant time steps that neural networks use for learning. 
The methodology proposed here consists of two steps (described in section~\ref{subsec:coarse-to-fine}), a coarse evaluation of the average performance, followed by the evaluation of robustness or uncertainty of these average performance. For the latter step, we devised quantitative  metrics to probe the score-drop distribution, that measure the robustness of post-hoc methods (described in section~\ref{subsec:fine-grained-eval}).

\subsection{Coarse-to-fine-grained level evaluation}\label{subsec:coarse-to-fine}
Coarse-grained evaluation aims to globally assess the ability of post-hoc interpretability methods in capturing relevant time steps in a dataset on average. To perform this evaluation step, we corrupt portions of the input time series associated to positive relevance, $\mathrm{R}^+$, yielding to a corrupted input $\bar{\mathrm{X}}$. In particular, we corrupt different numbers of time steps based on $\mathrm{R}^+$ k-percentile levels, where k represents the percentile level. Once the time series is corrupted for a certain k-percentile, the classifier produces a score, $\mathcal{S}(\bar{\mathrm{X}})$. For more details on the corruption strategy, see Appendix~\ref{subapp:corruption}. Once the k-percentile corruption is performed for all samples $\mathbf{X}$ in the testing set at different values of k, we can assess the coarse-grained performance of interpretability methods, looking at the average of normalized score drop $\tilde{\mathcal{S}}(\bar{\mathbf{X}})$ (see Appendix~\ref{subapp:corruption} for more details). To this end, we use the two metrics designed in~\cite{turbe2023evaluation}, namely $\operatorname{AUC}\tilde{\mathcal{S}}^\text{top}_\text{m}$ and $\operatorname{F1}\tilde{\mathcal{S}}_\text{m}$. The first measures the ability of post-hoc methods to identify the most relevant time steps -- see Appendix~\ref{subapp:coarse}. The second calculates an harmonic mean that evaluates the ability of the method to recognize both the most and least relevant time steps -- see Appendix~\ref{subapp:coarse}. We note that we use a slightly different notation from~\cite{turbe2023evaluation}, adding subscript $\mathrm{m}$ to emphasize that these metrics refer to mean (average) values. Indeed, these two metrics provide only a view of the average performance of post-hoc interpretability methods across an entire dataset.

\begin{figure}[t]
\begin{center}
\centerline{\includegraphics[width=0.7\textwidth]{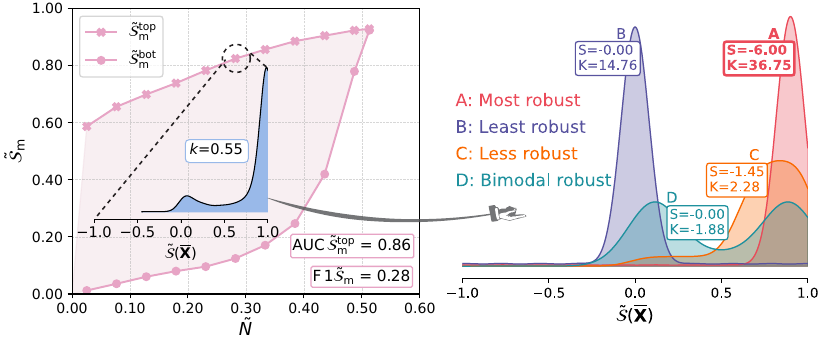}}
\caption{Left: $\tilde{\mathcal{S}}_\text{m}-\tilde{N}$ curve for coarse-grained evaluation and distribution example for fine-grained evaluation at top 55\% corruption. Right: four distribution shapes commonly observed to visualize the robustness of post-hoc interpretability methods.}
\label{fig:coarse_fine}
\end{center}
\end{figure}

To understand the robustness of post-hoc interpretability methods across samples we propose to compute the \textit{distribution} of the normalized score drop $\tilde{\mathcal{S}}(\mathcal{\overline{\mathbf{X}}})$ via kernel density estimation (KDE).
We perform this computation for each k-percentile corruption. 
In Figure \ref{fig:coarse_fine}, we show an example of the distribution obtained when corrupting the top 55\% time steps identified by DeepSHAP for a trained Transformer model on our synthetic dataset (for more details on the dataset see Appendix~\ref{subapp:dataset-synthetic}). 
The right plot of Figure \ref{fig:coarse_fine} depicts four different distributions that can be associated to the robustness of post-hoc interpretability methods. 
A distribution similar to shape A indicates that the post-hoc method is robust for the majority of samples in the testing set. 
In other words, the post-hoc method can correctly capture relevant time steps for most samples, and corrupting these time steps will substantially impact the prediction. 
Conversely, shape B shows the poorest robustness where the post-hoc method hardly identifies relevant information for most samples (it provide more uncertain results). 
Shape C suggests less robustness, with a relatively mediocre drop for some samples compared to shape A. 
The bimodal shape D presents polarized performance across all samples, capturing relevant information in some samples but failing in others. 
Using k-percentile corruption from 5\% to 95\%, with 10\% intervals, we obtain 10 different distributions of normalized score drops. 
We propose the use of ridge-line plots to visualize these distributions for a qualitative evaluation of post-hoc methods robustness.
A ridge-line plot example is depicted in Figure \ref{fig:ridge_skew_kurt}, accompanied by detailed illustrations and discussions in section~\ref{subsec:ridgeplot}. 
Around the distributions provided in the ridge-line plots, we can define quantitative metrics that can better capture the robustness of post-hoc interpretability methods, and that are introduced next.

\subsection{Fine-grained evaluation metrics}\label{subsec:fine-grained-eval}
To evaluate robustness, we adopt two statistical metrics, \textit{skewness} and \textit{kurtosis}, to quantitatively characterize the distribution shapes introduced in section~\ref{subsec:coarse-to-fine}. On the one hand, skewness measures the asymmetry of the score drop distribution, and helps quantifying how much the normalized score drop is skewed to a certain side. On the other hand, kurtosis quantifies the tailedness and peakedness of a score drop distribution compared to a normal distribution. 

We use sample skewness, which can be computed as the Fisher-Pearson coefficient of skewness:
\begin{equation}\label{eq:skew}
    \textit{skew}=\frac{\kappa_3}{{\kappa_2}^{3/2}}, \text{where} \  \kappa_i = \frac{1}{L}\sum^L_{n=1}(\tilde{S}(\bar{\mathrm{X}}_n)-\tilde{S}_{\mathrm{m}}^{\text {top }})^i,
\end{equation}
where, $\kappa_i$ is the $i$-th biased sample central moment, $\tilde{S}(\bar{\mathrm{X}}_n)$ is the normalized score drop of the $n^{th}$ time series sample, $\tilde{S}_{\mathrm{m}}^{\text {top }}$ is the mean normalized score drop under the top-k corruption strategy (i.e., when corrupting time steps from higher to lower k-percentile -- see also Appendix~\ref{subapp:corruption}) across all $L$ samples in the testing set. We use top-k as it provides information on whether post-hoc methods capture the most relevant time steps associated to the classification decision. The skewness coefficient, $\mathrm{skew}$, can be zero, negative, or positive. Symmetric distributions similar to Shape B and D (bimodal distribution) in Figure \ref{fig:coarse_fine} can both result in zero skewness. Negative skewness indicates a left-skewed (longer tail on the left side) distribution, while positive skewness indicates a right-skewed (longer tail on the right side) distribution. In this sense, the more negative the skewness, the more robust the post-hoc method is at each k-percentile corruption. 

For kurtosis, we use the excess kurtosis, which is defined as Pearson's kurtosis (standard measure of kurtosis) minus 3:
\begin{equation}\label{eq:kurt}
    \textit{(E)kurt}=\frac{\kappa_4}{{\kappa_2}^2} - 3.
\end{equation}
Excess kurtosis is a practically adopted measurement, allowing for easy comparison to the normal distribution with excess kurtosis of zero. Negative excess kurtosis represents a ``platykurtic" distribution (e.g., uniform distribution), while a positive excess kurtosis represents ``leptokurtic" distribution (e.g., Laplace distribution). In terms of post-hoc interpretability robustness, the higher the positive value of kurtosis the less spread the distribution. Hence, having a large positive value of kurtosis, with a large negative value of skewness means that a post-hoc method is robust across the dataset.

By calculating skewness and kurtosis of score drop distributions, as in \cref{eq:skew} and \cref{eq:kurt}, we can quantitatively assess the robustness of post-hoc interpretability methods. For instance, distribution A in Figure \ref{fig:coarse_fine} has a skewness of -6.00 and a kurtosis of 36.75, indicating its superior robustness (compared to the other distributions in the figure) in capturing informative time steps for most samples. 

In order to evaluate post-hoc interpretability methods, we are interested in the skewness and kurtosis results across different k-percentile corruptions. To this end, we rescale both metrics to $[0,1]$, and create $\textit{skew}$-$k$ and $\textit{(E)kurt}$-$k$ curves. One example of these two curvesis shown in Figure \ref{fig:ridge_skew_kurt}, top-right panel, obtained for the synthetic dataset (see Appendix~\ref{subapp:dataset-synthetic} for more details on the dataset). From these plots, we can compute the area under the $\textit{skew}$-$k$ and $\textit{(E)kurt}$-$k$ curves, and define two quantitative metrics, namely $\overline{\operatorname{AUC}\textit{Skew}}$ and $\operatorname{AUC}\textit{Kurt}$:
\begin{equation}
    \overline{\operatorname{AUC}\textit{Skew}} = 1 - \int_{0.05}^1 \textit{skew} \mathrm{~d} k,
    \label{eq:aucskew}
\end{equation}
\vskip -0.2in
\begin{equation}
    \operatorname{AUC}\textit{Kurt} = \int_{0.05}^1 \textit{(E)kurt} \mathrm{~d} k.
\end{equation}
Recall that a post-hoc method is more robust when the skewness is more negative and the kurtosis is more positive. Accordingly, a method with better robustness is expected to have smaller scaled skewness and larger scaled kurtosis. $\overline{\operatorname{AUC}\textit{Skew}}$ is designed to be the inverse of the area under the $\textit{skew}$-$k$ curve ($1-\operatorname{AUC}\textit{Skew}$) in \cref{eq:aucskew}. Thereby, regarding fine-grained quantitative evaluation, a post-hoc interpretability method is considered more robust when the two metrics, $\overline{\operatorname{AUC}\textit{Skew}}$ and $\operatorname{AUC}\textit{Kurt}$, are larger.
\begin{figure*}[t]
\begin{center}
\centerline{\includegraphics[width=\textwidth]{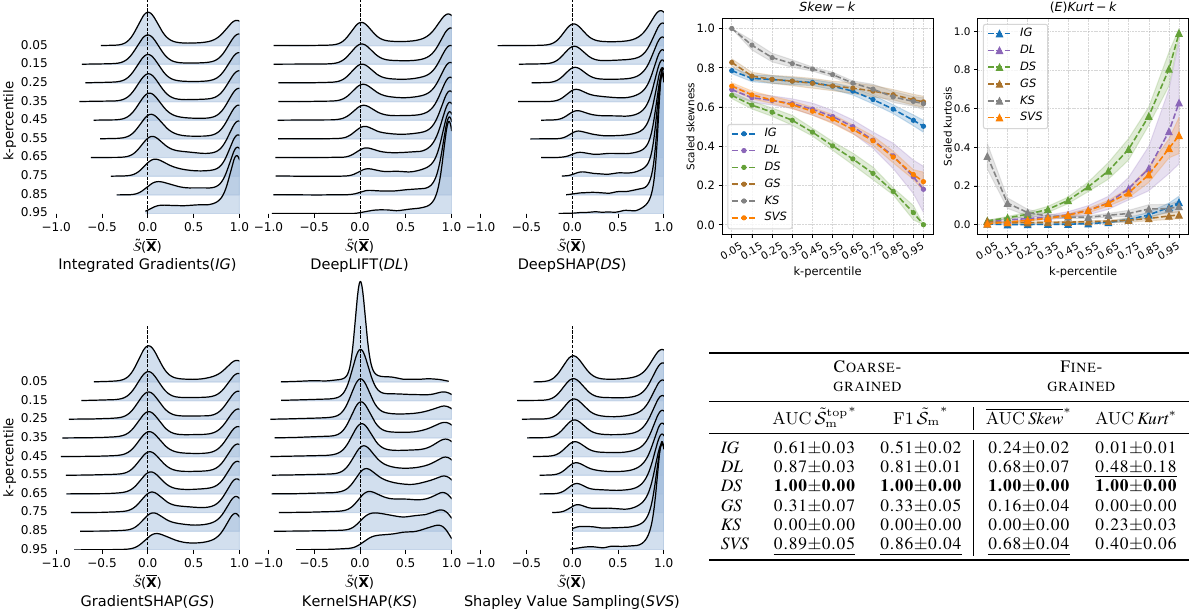}}
\caption{Left: ridge-line visualization of six post-hoc interpretability methods for a trained Transformer model on synthetic dataset. Top-right: scaled $\textit{skew}$-$k$ curve and $\textit{(E)kurt}$-$k$ curve. Bottom-right: coarse-grained and fine-grained metrics of six methods.}
\label{fig:ridge_skew_kurt}
\end{center}
\vskip -0.4in
\end{figure*}
%

\section{Results} 
\label{sec:4}

\textbf{Datasets.} We apply our evaluation framework to one synthetic dataset and 20 public datasets. The synthetic dataset is composed of four features, and was first proposed in~\cite{turbe2023evaluation}. The 20 public datasets include the ECG database from CPSC~\cite{alday2020classification}, and other 19 datasets from the \href{https://www.timeseriesclassification.com/}{UCR time series classification archive}~\cite{dau2019ucr}. Dataset details can be found in Appendix~\ref{app:dataset}.

\textbf{Neural network architectures.} We implement our evaluation framework for CNN, BiLSTM, and Transformer architectures, to comprehensively investigate statistical results across different neural networks. See Appendix~\ref{app:neuralnets} for details.

\textbf{Post-hoc interpretability methods.} We consider six commonly used post-hoc methods implemented in the \href{https://captum.ai}{Captum library} \cite{kokhlikyan2020captum}: \textit{Integrated Gradients (IG)}~\cite{sundararajan2017axiomatic}, \textit{DeepLIFT (DL)}~\cite{shrikumar2017learning}, \textit{DeepSHAP (DS)}~\cite{lundberg2017unified}, \textit{GradientSHAP (GS)}~\cite{lundberg2017unified}, \textit{KernelSHAP (KS)}~\cite{lundberg2017unified}, and \textit{Shapley Value Sampling (SVS)}~\cite{castro2009polynomial}.

\subsection{Qualitative evaluation: Ridge-line visualization} \label{subsec:ridgeplot}
We consider the synthetic dataset case as an example to illustrate the proposed ridge-line visualization for a qualitative fine-grained post-hoc interpretability evaluation. 
In Figure \ref{fig:ridge_skew_kurt}, we show the ridge-line plots (left panel) of six post-hoc interpretability methods for the trained Transformer model applied to the synthetic dataset. 
The Transformer model achieves 92\% accuracy on the testing set, composed of 1500 samples. 
Six post-hoc methods are deployed to generate relevance scores for the testing set. 
For each top-k percentile corruption from 5\% to 95\%, we show the score drop distribution, obtained through the KDE of the normalized score drop $\tilde{\mathcal{S}}(\mathcal{\bar{\mathbf{X}}})$ for all 1500 samples. 
By visually inspecting the ridge-line plots, we can grasp a qualitative understanding of post-hoc methods robustness. 
A post-hoc method has better robustness if its distribution is left-skewed (negative skewness) and ``leptokurtic" (positive kurtosis), starting at a relatively low k-percentile corruption. For instance, the ridge-line plot of \textit{DS} in Figure \ref{fig:ridge_skew_kurt}, left panel, shows that for small k-percentile value, the distribution shifts significantly to the right, thereby showing that \textit{DS} is able to capture the most informative time steps used by the classifier to make its decision across all samples in the testing set.

From Figure \ref{fig:ridge_skew_kurt}, left panel, we can infer that \textit{DS}, \textit{DL} and \textit{SVS} exhibit similar and superior robustness than e.g., \textit{KS}, which demonstrates the poorest robustness among the six interpretability methods. These results are confirmed by the $\textit{skew}$-$k$ and $\textit{(E)kurt}$-$k$ curves shown in Figure \ref{fig:ridge_skew_kurt}, top-right panel, where the curves associated to the three methods are the lowest and highest for skewness and kurtosis, respectively. The peak of the  distribution is located near 1.0 for all three methods, indicating that they identify relevant time steps for the majority of samples in the testing set. If we look at the worst performing method, namely \textit{KS}, the score drop distribution shows that it can hardly identify informative points for almost all samples. 

As mentioned in section~\ref{sec:intro}, the drop in score might also depend on the sensitivity of certain samples to input corruption. In particular, samples close to the decision boundary of the classifier are considered sensitive, and the model might be uncertain about these samples. In contrast, samples far from the decision boundary might prove less sensitive. To study the sensitivity to the decision boundary, we calibrate the Transformer model during training for the synthetic dataset. We present the results in Appendix~\ref{app:calibration}, and show that calibration can improve the robustness of post-hoc interpretability methods, but leaves comparative performance across interpretability methods unchanged. This means that the evaluation framework proposed is effective in ranking post-hoc interpretability methods performance regardless of sample sensitivity to corruption.

\begin{figure*}[t]
\centering
\captionsetup[subfloat]{captionskip=-0.05in}
\subfloat[Coarse-grained metrics]{\label{fig:boxplots_coarse}\includegraphics[width=\textwidth]{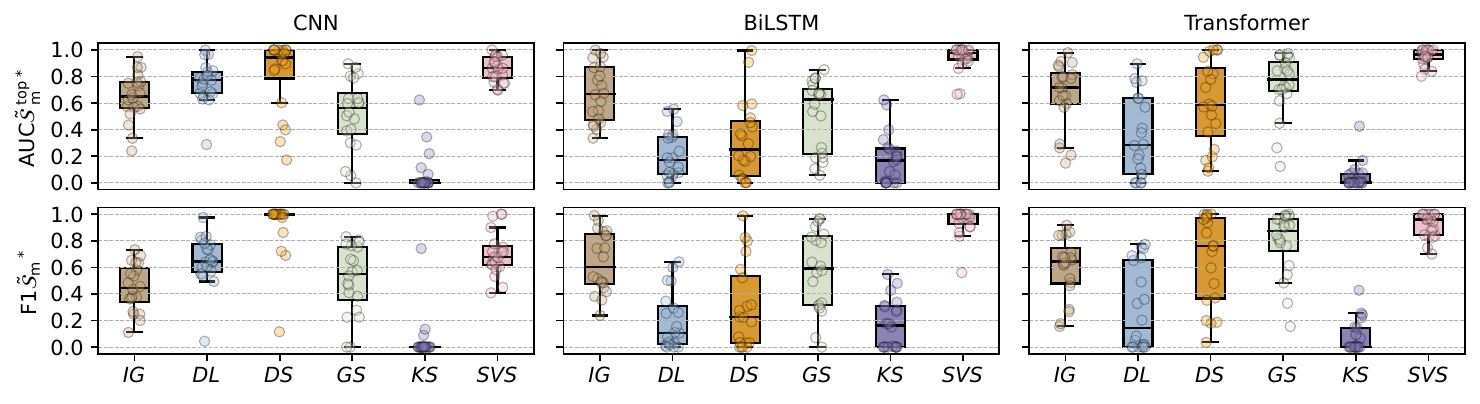}}\\
\vspace{-0.15in}
\subfloat[Fine-grained metrics]{\label{fig:boxplots_fine}\includegraphics[width=\textwidth]{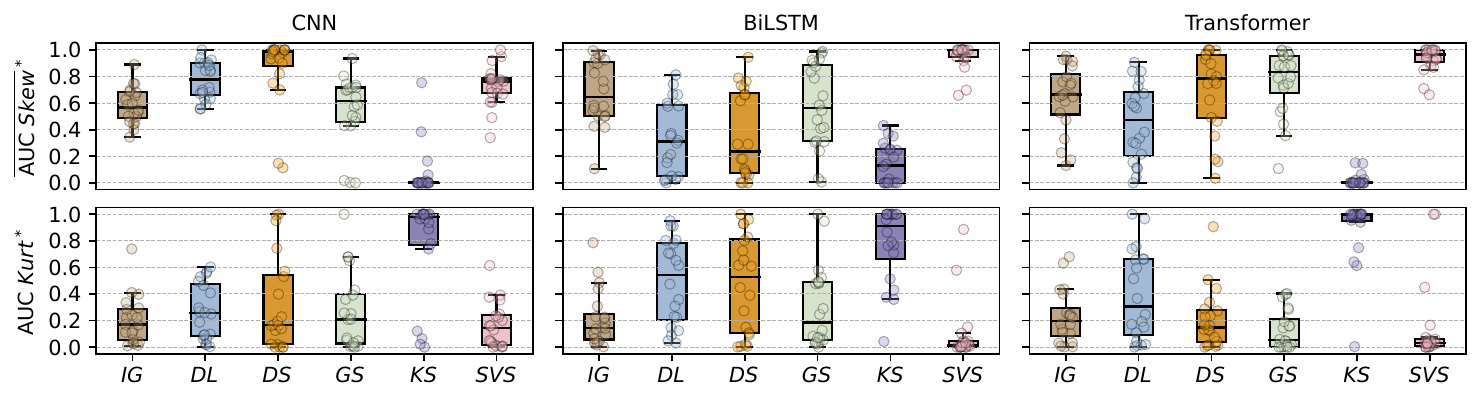}}
\caption{Statistical coarse-grained and fine-grained metrics for CNN, BiLSTM, and Transformer models across 20 public datasets.}
\vskip -0.2in
\end{figure*}

\subsection{Quantitative evaluation: $\overline{\text{AUC}\textit{Skew}}^*$ and {AUC}\textit{Kurt}$^*$} 
\label{subsec: metrics}
We consider the synthetic dataset to illustrate the use of the fine-grained quantitative metrics proposed, namely $\overline{\text{AUC}\textit{Skew}}^*$ and {AUC}\textit{Kurt}$^*$. Figure \ref{fig:ridge_skew_kurt}, bottom-right panel (table), shows the two fine-grained metrics for the Transformer model trained on the synthetic dataset and for all six post-hoc interpretability methods. In the same table we also show the coarse-grained metrics, $\operatorname{AUC}\tilde{\mathcal{S}}^\text{top}_\text{m}$ and $\operatorname{F1}\tilde{\mathcal{S}}_\text{m}$. For each experiment, we perform 5 repetitions, to get the mean and standard deviation of all metrics shown. We also rescale the metrics to $[0,1]$ across the six post-hoc methods for a comparative analysis. The standardized metrics are marked with superscript $(^*)$. The table in Figure \ref{fig:ridge_skew_kurt}, bottom-right panel, shows that \textit{DS} is the best performing method for both coarse-grained and fine-grained metrics. This means that it provides the best coarse-grained (average) interpretations as well as the most robust ones (fine-grained).
The second and third best performing methods in terms of coarse-grained metrics, namely \textit{SVS} and \textit{DL}, follow closely \textit{DS}. Yet, when it comes to fine grained metrics, they seem to fall well behind \textit{DS}, with scores that are significantly lower. Indeed, if we inspect the ridge-line plots (left panel in Figure \ref{fig:ridge_skew_kurt}), we observe that the distributions associated to \textit{SVS} and \textit{DL} start becoming skewed toward a score drop of 1 for higher k-percentile, as highlighted by the top-right panel in Figure \ref{fig:ridge_skew_kurt}. 
We also note that \textit{KS} provides the worst coarse- and fine-grained interpretations. 

The quantitative assessment through coarse- and fine-grained metrics allows selecting the post-hoc interpretability method that performs best for the task at hand, in terms of average performance, and robustness (i.e., uncertainty). We note that robustness is generally linked to coarse-grained performance, yet, there can be differences in terms of methods ranking using coarse- vs fine-grained metrics. Depending on the specific application, the post-hoc method selection may not be trivial, and could value average performance vs robustness differently. In addition, practitioners may also take into account the computational costs associated to the interpretability methods, that may play a factor in the selection procedure.

\subsection{Sensitivity to dataset and model}
\label{subsec:statistical}
To thoroughly test the proposed evaluation framework, we use 3 different neural network architectures and 20 public datasets. Figure \ref{fig:boxplots_coarse} and \ref{fig:boxplots_fine} show the the coarse-grained and fine-grained metrics of the six post-hoc interpretability methods (x-axis of each plot), where each plot-column represents a different neural network, i.e., the left plots depict CNN, the middle plots BiLSTM, and the right plots Transformer. For each post-hoc interpretability method applied to a neural network architecture, we display metrics of 20 datasets using circle dots. Additionally, box-plots  visualize the statistical distribution of different post-hoc interpretability methods. 

\textit{DS} is the best (coarse-grained) and the most robust (fine-grained) method for CNN. \textit{SVS} consistently performs best, both coarse- and fine-grained for BiLSTM and Transformers models. For BiLSTM and Transformers \textit{IG}, \textit{DL}, \textit{DS}, and \textit{GS} have longer boxes, which means these methods have a larger interquartile range of metrics. It indicates these four methods successfully identify important information used by BiLSTM and Transformer models in some datasets but fail in others. As shown in Figure \ref{fig:boxplots_coarse}, \textit{KS} ranks last in all datasets across all three neural network architectures, indicating that it is unable to capture essential time steps that neural networks use for the learning task. Overall, \textit{SVS} provide reliable results for the majority of architectures and datasets, both in terms of coarse- and fine-grained performance.

\subsection{Sensitivity to noise}
\label{subsec:noise}
We investigate the sensitivity of the six interpretability methods to noise. We keep the same hyperparameters of CNN, BiLSTM, and Transformer models trained on the synthetic dataset without noise, and train corresponding models for the dataset with three different amplitudes of Gaussian noise. The signal-to-noise ratio of these three versions of noise are, $\mathrm{SNR_{dB}} = 20\mathrm{dB}, 15\mathrm{dB}, 10\mathrm{dB}$. The detailed results can be found in Appendix~\ref{app:sensnoise}. Table \ref{table:syn_noise} shows the sensitivity to noise of different neural network architectures and post-hoc methods. Note that \textit{KS} ranks last among six methods based on coarse-grained metrics, and its high value of ${\operatorname{AUC}\textit{Kurt}}^*$ is caused by the peak shape of score drop distribution around zero (that means poor fine-grained performance). After introducing noise in the synthetic dataset, \textit{SVS} is the most robust post-hoc interpretability method for interpreting different classifiers, according to both coarse- and fine-grained metrics presented in \cref{table:syn_noise}.



\section{Concluding remarks}
\label{sec:5}

We proposed a new way of thinking about the robustness and evaluation of post-hoc interpretability methods. In particular, we provided a new framework, namely coarse-to-fine-grained evaluation, as well as new metrics to evaluate the robustness of post-hoc interpretability methods. This entails understanding both the average performance of interpretability methods through a coarse-grained evaluation, and the uncertainty associated to these average performance through a fine-grained evaluation. The latter aspect is achieved by looking at the properties of the statistical distributions associated to the score drops obtained after corrupting relevant inputs. To this end, we developed two metrics that measure the skewness and the kurtosis of the distribution across different k-percentile corruptions, and allow for a quantitative understanding of the robustness (or uncertainty) of the coarse-grained evaluation. 

We tested the framework and metrics across 20 different public datasets, as well as 1 synthetic dataset for 3 different neural network architectures. The results show consistent results across neural networks and datasets, and are able to rank the methods in terms of average performance and robustness. The two aspects are frequently related, albeit not always, and practitioners may value average performance vs robustness differently depending of the specific applications. Yet, our framework allows for quantifying these two key aspects, and can make stakeholders, especially in critical application areas, more prone to use deep learning solutions. Our work  is in line with increasing regulations of machine learning, and demand for transparency~\cite{ai_act}.

We provide the code used to fully reproduce all the results presented in this paper. However, due to size, we are unable to upload the datasets. We provide the code and steps to generate the synthetic dataset, and the instructions to download and format all the 20 public datasets. Refer to Appendix~\ref{app:reproducibility}.

\section{Broader impact and ethical considerations}\label{sec:6}
This work aims to improve the transparency of deep learning models, focusing on robustness, a key aspect for trustworthy machine learning. We do not envision any negative ethical implications of the current work, while we expect positive future societal outcomes, as we address trust issues towards deep learning models. 

\bibliographystyle{unsrtnat}
\bibliography{references}

\newpage
\appendix
\onecolumn
\setcounter{equation}{0}
\renewcommand{\theequation}{\thesubsection. \arabic{equation}}
\section{Details of methodology}\label{app:methods}

\subsection{Corruption-based evaluation}\label{subapp:corruption}
Assuming that time steps with higher relevance scores are more important than others, we sort the positive relevance $\mathrm{R}^+$ to form an ordered set $\{\mathrm{R}_{e}=(r_{m,t})\}_{e=1}^{{[M \times T]}^+}$, where $\mathrm{R_e}$ is the $e^{th}$ element and ${[M \times T]}^+$ is the total number of points with positive relevance. Given that different post-hoc methods may recognize varying numbers of positive scores for one sample, we use k-percentile corruption rather than corrupting a constant number of time steps:
\begin{equation}
    P_k = [M \times T]^+ \times k, \text{where} \  k \in \{0.05, 0.15, ..., 1.00\}
\end{equation}
Meanwhile, to comprehensively understand the capability of post-hoc methods to identify the most and least relevant points, we conduct top-k and bot-k corruption strategies. Specifically, we sort the positive relevance in a descending order $\{\mathrm{R}^{\text{top}}_{d \downarrow}\}_{d=1}^{{[M \times T]}^+}$ for the top-k corruption schema, while ascending order $\{\mathrm{R}^{\text{bot}}_{a \uparrow}\}_{a=1}^{{[M \times T]}^+}$ for the bot-k schema. For each time series sample, we replace the first k\% points with the points drawn from a Gaussian distribution, for both top-k and bot-k corruption schemes:
\begin{equation}
\bar{\mathrm{X}}_k^{\mathrm{top}}:= \begin{cases}\mathcal{N}(0,1) & \text { if } r_{m, t} \in\left\{\mathrm{R}_{d \downarrow}^{\mathrm{top}}\right\}_{d=1}^{P_k} \\ x_{m, t} & \text { otherwise }\end{cases}
\end{equation}
\begin{equation}
\bar{\mathrm{X}}_k^{\mathrm{bot}}:= \begin{cases}\mathcal{N}(0,1) & \text { if } r_{m, t} \in\left\{\mathrm{R}_{a \uparrow}^{\mathrm{bot}}\right\}_{a=1}^{P_k} \\ x_{m, t} & \text { otherwise }\end{cases}
\end{equation}

Given two different inputs, $\mathrm{X}$ and $\mathcal{\bar{\mathrm{X}}}$, the trained neural network has two different outputs, $\mathcal{S}(\mathrm{X})$ and $\mathcal{S}(\mathcal{\bar{\mathrm{X}}})$. Corrupting time steps can result in significant differences in the output of the neural network if one post-hoc method identifies the necessary time steps for a correct prediction. We introduce the normalized score drop to build quantitative metrics:
\begin{equation}
    \tilde{\mathcal{S}}(\mathcal{\bar{\mathrm{X}}}) = \frac{\mathcal{S}(\mathrm{X}) - \mathcal{S}(\mathcal{\bar{\mathrm{X}}})}{\mathcal{S}(\mathrm{X})}
\end{equation}
Interpretability evaluation in this work conduct on testing set $[\mathbb{T}]$, thus all quantitative metrics aim to analyze the normalized score drop $\tilde{\mathcal{S}}(\mathcal{\bar{\mathbf{X}}})$ for all testing samples $\mathbf{X}$.

\setcounter{equation}{0}
\subsection{Coarse-grained metrics}\label{subapp:coarse}
After implementing top-k and bot-k corruption, we can calculate the mean score drop $\tilde{\mathcal{S}}^\text{top}_\text{m}(\mathcal{\bar{\mathbf{X}}})$ and $\tilde{\mathcal{S}}^\text{bot}_\text{m}(\mathcal{\bar{\mathbf{X}}})$ at each k-percentile corruption, respectively. Let $N=M \times T$ represent the total number of time
steps for each sample, we obtain the mean ratio of corruption $\tilde{N} = \frac{\bar{N}}{N}$, where $\bar{N}$ is the mean number of time steps corrupted across all samples.
This in mind, it is possible to build $\tilde{\mathcal{S}}^\text{top}_\text{m}-\tilde{N}$ and $\tilde{\mathcal{S}}^\text{bot}_\text{m}-\tilde{N}$ curves to track the mean normalized score drop as corruption increases, as shown in the example at the top of Figure \ref{fig:coarse_fine}.

With $\tilde{\mathcal{S}}_\text{m}-\tilde{N}$ curve, we adopt two metrics from \cite{turbe2023evaluation} as coarse-grained evaluation, to assess the capability of post-hoc interpretability methods in relevance identification at the dataset level:
\begin{equation}
\operatorname{AUC} \tilde{\mathcal{S}}_{\mathrm{m}}^{\text {top }}=\int_0^1 \tilde{\mathcal{S}}_{\mathrm{m}}^{\text {top }} \mathrm{d} \tilde{N}
\end{equation}
\begin{equation}
\mathrm{~F} 1 \tilde{\mathcal{S}}_{\mathrm{m}}=\frac{\operatorname{AUC} \tilde{\mathcal{S}}_{\mathrm{m}}^{\text {top }}\left(1-\mathrm{AUC} \tilde{\mathcal{S}}_{\mathrm{m}}^{\text {bot }}\right)}{\operatorname{AUC} \tilde{\mathcal{S}}_{\mathrm{m}}^{\text {top }}+\left(1-\mathrm{AUC} \tilde{\mathcal{S}}_{\mathrm{m}}^{\text {bot }}\right)}
\end{equation}

\setcounter{table}{0}
\renewcommand{\thetable}{\thesection. \Roman{table}}
\setcounter{figure}{0}
\renewcommand{\thefigure}{\thesection. \Roman{figure}}
\section{Dataset.}\label{app:dataset}
\subsection{Synthetic dataset}\label{subapp:dataset-synthetic}
The synthetic dataset is inspired from \cite{turbe2023evaluation}, compositing four sine waves (500 time steps) with baseline frequency sampled from uniform distribution $U(2, 5)$. Two supported sine blocks (100 time steps) with higher frequencies $f_1$ and $f_2$ sampled from $ U(10, 50)$ are randomly added to the base waves. Thereby, the sum of two supported frequencies is range from 20 to 100, i.e., $(f_1 + f_2) \in [20, 100]$. To make it a balanced dataset, we setup a threshold of 60. The classification task is to predict whether the sum $(f_1 + f_2)$ is above or below the threshold.


\subsection{Public datasets}\label{subapp:dataset-public}
The information of 20 public time series classification datasets for experimental results shown in this paper can be found in \cref{table:PubDataInfo}, and they are from the UCR database (19) and the CPSC dataset (1). The 20 datasets cover different problems, including  i) univariate and multivariate series; ii) binary and multiclass tasks; iii) big and small data sizes. The 19 UCR datasets we used can be found at \url{https://www.timeseriesclassification.com/}, while the CPSC dataset can be found at \url{https://physionet.org/content/challenge-2020/1.0.2/}.
\begin{table}[h]
\caption{20 public datasets listing, comprehensively including univariate/multivariate and binary/multiclass classification tasks.}
\label{table:PubDataInfo}
\begin{center}
\begin{small}
\begin{sc}
\begin{tabular}{lcccc}
\toprule
Dataset Name & Size & Length & No. of Features &  No. of Classes \\
\midrule
Blink & 950 & 510 & 4 & 2 \\
CBF & 930 & 128 & 1 & 3 \\
ECG & 6877 & 450 & 12 & 2 \\
ElectricDevices & 16637 & 96 & 1 & 7 \\
EMOPain & 1143 & 180 & 30 & 3 \\
FordA & 4921 & 500 & 1 & 2 \\
FreezerRegularTrain & 3000 & 301 & 1 & 2 \\
GunPointAgeSpan & 451 & 150 & 1 & 2 \\
ItalyPowerDemand & 1096 & 24 & 1 & 2 \\
MoteStrain & 1272 & 84 & 1 & 2 \\
MotionSenseHAR & 361 & 200 & 12 & 6 \\
PenDigits & 10992 & 8 & 2 & 10 \\
SonyAIBORobotSurface1 & 621 & 70 & 1 & 2 \\
SonyAIBORobotSurface2 & 980 & 65 & 1 & 2 \\
SpokenArabicDigits & 8798 & 93 & 13 & 10 \\
StarLightCurves & 9236 & 1024 & 1 & 3 \\
Strawberry & 983 & 235 & 1 & 2 \\
TwoLeadECG & 1162 & 82 & 1 & 2 \\
TwoPatterns & 5000 & 128 & 1 & 4 \\
Wafer & 7164 & 152 & 1 & 2 \\
\bottomrule
\end{tabular}
\end{sc}
\end{small}
\end{center}
\vskip -0.1in
\end{table}

\setcounter{table}{0}
\renewcommand{\thetable}{\thesection. \Roman{table}}
\setcounter{figure}{0}
\renewcommand{\thefigure}{\thesection. \Roman{figure}}
\section{Neural network architectures.}\label{app:neuralnets}
For all datasets mentioned in this paper, we implement three neural network architectures for training, CNN, BiLSTM, and Transformer. In this appendix, we provide model details for the synthetic dataset case. Configurations for the 20 public datasets will be made available on GitHub upon acceptance.
The configurations of trained models and details of hyperparameter optimization on the synthetic dataset are provided below.

\textbf{Configurations of trained CNN model on the synthetic dataset}
\begin{itemize}[itemsep=0pt]
    \item Layer1: Conv 1D [Units=128, Kernelsize=11, Stride=1, Dropout=0.2]
    \item Layer2: Conv 1D [Units=128, Kernelsize=11, Stride=1, Dropout=0.2]
    \item Layer3: Conv 1D [Units=128, Kernelsize=11, Stride=1, Dropout=0.2]
    \item Dense layer: Units=128
    \item Activation function: ReLU
    \item Optimizer: RAdam
    \item Learning rate: 0.001
    \item Batch size: 64
\end{itemize}

\textbf{Configurations of trained BiLSTM model on the synthetic dataset}
\begin{itemize}[itemsep=0pt]
    \item LSTM1: BiLSTM units=64, Dense units=64
    \item LSTM2: BiLSTM units=128, Dense units=128
    \item LSTM3: BiLSTM units=64, Dense units=64
    \item Dense layer: Units=256
    \item Activation function: ReLU
    \item Optimizer: RAdam
    \item Learning rate: 0.001
    \item Batch size: 64
\end{itemize}

\textbf{Configurations of trained Transformer model on the synthetic dataset}
\begin{itemize}[itemsep=0pt]
    \item Layer1: Conv 1D [units=128, kernelsize=3, stride=1, dropout=0.2]
    \item Layer2: Conv 1D [units=64, kernelsize=3, stride=1, dropout=0.2]
    \item Transformer layer1: heads=4, feedforward dim=256, dropout=0, mlp dim=64
    \item Transformer layer2: heads=4, feedforward dim=256, dropout=0, mlp dim=64
    \item Transformer layer3: heads=4, feedforward dim=256, dropout=0, mlp dim=64
    \item Dense layer: Units=64
    \item Activation function: ReLU
    \item Optimizer: RAdam
    \item Learning rate: 0.002
    \item Weight decay: 0.001
    \item Batch size: 64
\end{itemize}

\textbf{Hyperparameter Optimization}\\

We optimize hyperparameters for training models if the testing accuracy is below 90\%. We use  \href{https://hydra.cc/docs/1.1/plugins/optuna_sweeper/#internaldocs-banner}{\textit{Optuna Sweeper}} plugin in Hydra~\cite{Yadan2019Hydra} for hyperparameter search. For parameter sampling, we use the \href{https://optuna.readthedocs.io/en/stable/reference/samplers/generated/optuna.samplers.TPESampler.html#optuna.samplers.TPESampler}{TPESampler} provided by Optuna, which uses TPE (Tree-structured Parzen Estimator) algorithm.
Hyperparameter search space is provided as follows:
\begin{table}[ht]
\caption{Hyperparameter optimization for synthetic dataset.}
\label{table:hyop}
\begin{center}
\begin{small}
\begin{sc}
\begin{tabular}{lccc}
\toprule
& Hyperparameters & Search space & Sampling \\
\midrule
\multirow{4}{*}{Common}  & Learning rate & [0.0001, 0.1] & Interval \\
& Weight decay & [0, 0.001, 0.002] & Choice \\
& Dropout & [0, 0.1, 0.2, 0.3, 0.4, 0.5] & Choice \\
& Batch size & [32, 64, 128] & Choice \\
\midrule
\multirow{3}{*}{CNN}  &  No. of layers & [3, 4] & Choice\\
& No. of units & [32, 64, 128. 256] & Choice \\
& Kernel size & [3, 5, 7, 11, 15] & Choice \\
\midrule
\multirow{3}{*}{BiLSTM} & No. of layers & [3, 4] & Choice\\
& No. of units & [32, 64, 128. 256] & Choice \\
\midrule
\multirow{4}{*}{Transformer} & No. of layers & [2, 3, 4] & Choice\\
& No. of heads & [2, 4, 8] & Choice \\
& Feedforward dim & [32, 64, 128, 256] & Choice \\
& MLP dim & [32, 64, 128, 256] & Choice \\
\bottomrule
\end{tabular}
\end{sc}
\end{small}
\end{center}
\vskip -0.1in
\end{table}
\setcounter{figure}{0}
\renewcommand{\thefigure}{\thesection. \Roman{figure}}
\setcounter{equation}{0}
\renewcommand{\theequation}{\thesection. \arabic{equation}}
\section{Results of calibrated model on synthetic dataset}\label{app:calibration}

\begin{figure*}[ht]
\centering
\subfloat[Class 0: $(f_1 + f_2)<60$]{\label{fig:calibration1}\includegraphics[width=0.57\textwidth]{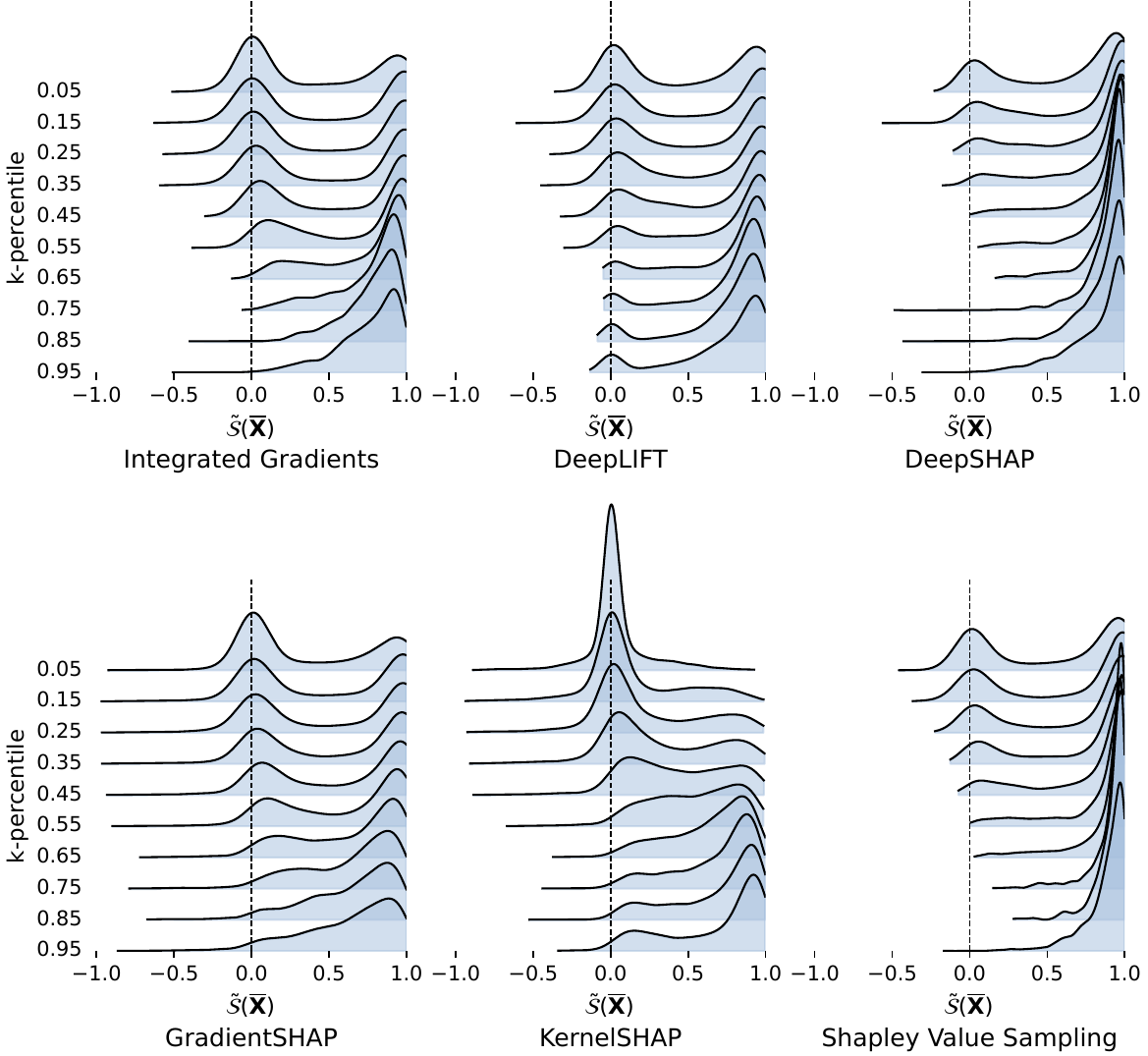}}
\hspace{0.1in}
\subfloat[Class 1: $(f_1 + f_2) \geq 60$]{\label{fig:calibration2}\includegraphics[width=0.57\textwidth]{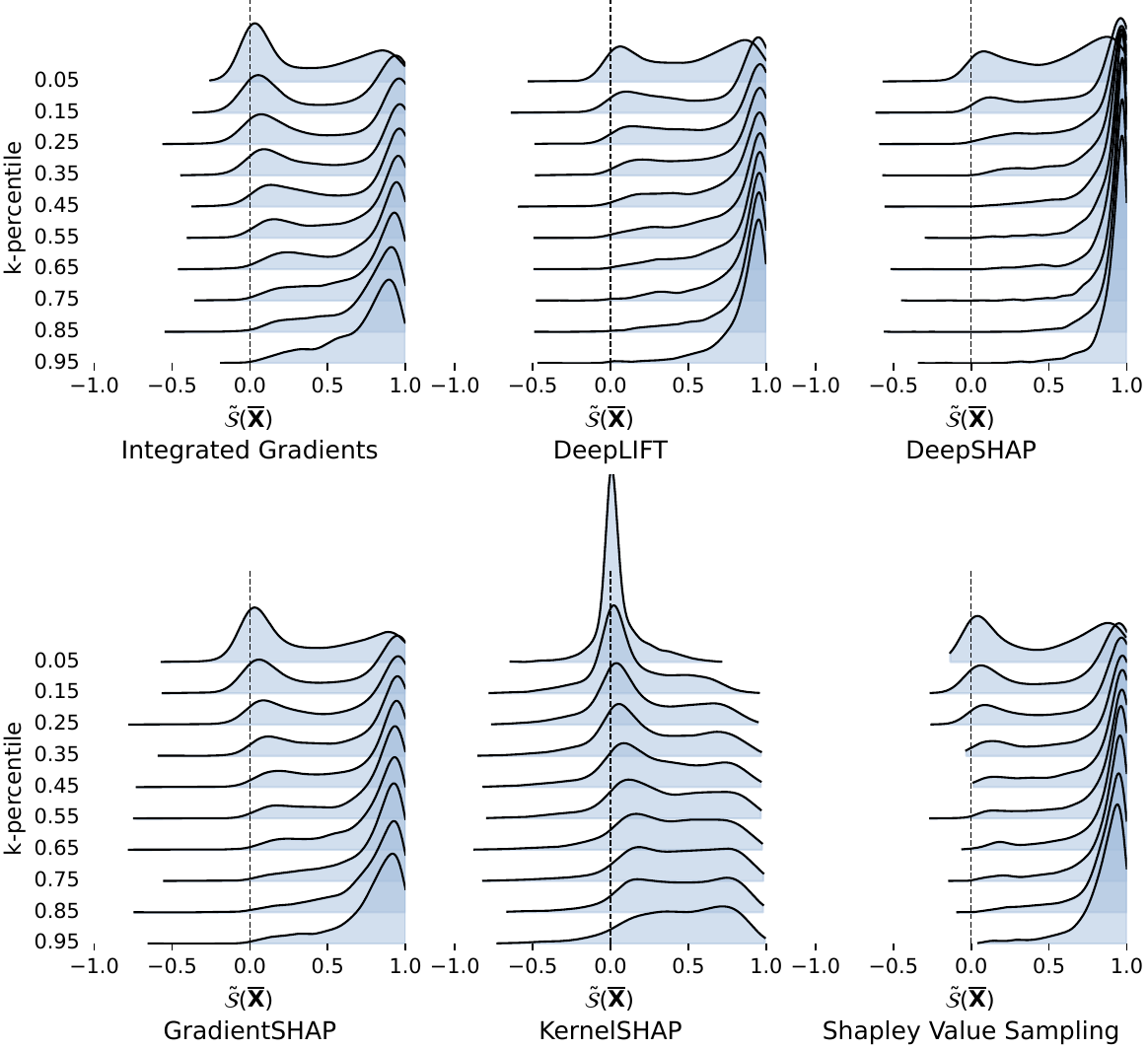}}
\caption{Ridge-line visualization of Calibrated Transformer based on Class 0 and Class 1.}
\end{figure*}
\textbf{Model calibration loss design}\\

With the information of $(f_1 + f_2)$ for each sample in the synthetic dataset, we can calibrate the predicted probabilities of the neural network by incorporating the frequency information into the loss function. Specifically, the calibrated loss function consists of two components: 1) cross entropy (CE) loss for the classification task; 2) mean squared error (MSE) loss for calibrating probabilities by embedding the frequency information, i.e., distance to the decision boundary.

As the synthetic case is a binary classification, we are able to calibrate the model based on either of the two classes (Class 0: $(f_1 + f_2)<60$; Class 1: $(f_1 + f_2) \geq 60$). First, we normalize the sum $(f_1 + f_2) \in [20, 100]$ to the range of $f_\mathrm{norm} \in [0, 1]$. We assume that the neural network will be more sensitive to samples near the decision boundary. The model may have lower confidence in the probabilities of these samples, which are expected to be around 0.5 rather than 1.0. Base on this, it is possible for us to design the calibrated loss choosing Class 0:
\begin{equation}
    Loss = CEloss + MSEloss(Prob[: \mathbf{0}], 1-f_\mathrm{norm})
\end{equation}
Or choosing Class 1:
\begin{equation}
    Loss = CEloss + MSEloss(Prob[: \mathbf{1}], f_\mathrm{norm})
\end{equation}
Taking the sample $(f_1 + f_2) = 40$ which belongs to Class 0 as an example, we could explain how we calibrate the model. After the normalization, $f_\mathrm{norm}$ of this sample should be 0.250. If we calibrate the model based on Class 0, we would expect the predicted probability of this sample to be close to $1-f_\mathrm{norm}$, that is, 0.750. Thus, the probability might reflect the distance from the threshold, of which the probability is approximately 0.5. On the other hand, if we choose Class 1 to calibrate the model, the expected probability of this sample would be $f_\mathrm{norm}$ itself, that is, 0.250.

Given the model calibration loss function, we keep the hyperparameters mentioned in Appendix~\ref{app:neuralnets} to train the calibrated model. Figure \ref{fig:calibration1} and \ref{fig:calibration2} shows the ridge-line plots of calibrated Transformer model based on Class 0 and 1, respectively. Compared to the ridge-line plot of the Transformer model without calibration in Figure \ref{fig:ridge_skew_kurt}, we observe the improvement significantly. Distributions in Figure \ref{fig:calibration1} and \ref{fig:calibration2} have a more left-skewed (longer tail on the left side) and ``leptokurtic" shape. It means that, calibrating the neural network could improve the robustness of post-hoc interpretability methods. This also indicates that our evaluation framework effectively and comprehensively assess different post-hoc interpretability methods. Meanwhile, it could provide useful information to understand and improve the model performance.


\setcounter{table}{0}
\section{Sensitivity of post-hoc interpretability methods to noise}\label{app:sensnoise}
For the synthetic case, we also investigate whether different post-hoc interpretability methods robust to noise in the data. \cref{table:syn_noise} provides coarse-grained and fine-grained metrics for four synthetic cases: 1) without noise; 2) with $\mathrm{SNR_{dB}} = 20\mathrm{dB}$ noise; 3) with $\mathrm{SNR_{dB}} = 15\mathrm{dB}$ noise; and 4) with $\mathrm{SNR_{dB}} = 10\mathrm{dB}$ noise. We perform 5 repetitions for each case, and finally get the mean and standard deviation of all metrics across these 4 cases.
\begin{table*}[ht]
\caption{Average coarse-grained and fine-grained metrics for synthetic dataset with three versions of Gaussian noise added to it.}
\label{table:syn_noise}
\begin{center}
\begin{small}
\begin{sc}
\begin{tabular}{lccccccc}
\toprule
Metrics & Networks & {\textit{IG}} & {\textit{DL}} & {\textit{DS}} & {\textit{GS}} & {\textit{KS}} & {\textit{SVS}}\\ 
\midrule
\multirow{3}{*}{${\operatorname{AUC}\tilde{\mathcal{S}}^\mathrm{top}_\mathrm{m}}^*$} & CNN & 0.73$\pm$0.07 & 0.72$\pm$0.11 & \underline{0.95$\pm$0.05} & 0.73$\pm$0.06 & 0.00$\pm$0.00 & \textbf{0.99$\pm$0.02} \\
& BiLSTM & \underline{0.72$\pm$0.03} & 0.31$\pm$0.10 & 0.38$\pm$0.12 & 0.49$\pm$0.05 & 0.00$\pm$0.00 & \textbf{1.00$\pm$0.00} \\
& Trans & 0.74$\pm$0.13 & 0.75$\pm$0.16 & \underline{0.84$\pm$0.25} & 0.64$\pm$0.21 & 0.00$\pm$0.00 & \textbf{0.95$\pm$0.06} \\ 
\midrule
\multirow{3}{*}{${\operatorname{F1}\tilde{\mathcal{S}}_\mathrm{m}}^*$} & CNN & 0.55$\pm$0.03 & 0.61$\pm$0.09 & \textbf{1.00$\pm$0.00} & 0.52$\pm$0.03 & 0.00$\pm$0.00 & \underline{0.74$\pm$0.03}\\
& BiLSTM & \underline{0.72$\pm$0.04} & 0.31$\pm$0.09 & 0.39$\pm$0.11 & 0.49$\pm$0.04 & 0.00$\pm$0.00 & \textbf{1.00$\pm$0.00} \\
& Trans & 0.74$\pm$0.15 & 0.73$\pm$0.14 & \underline{0.81$\pm$0.25} & 0.61$\pm$0.20 & 0.00$\pm$0.00 & \textbf{0.97$\pm$0.06} \\ 
\midrule
\midrule
\multirow{3}{*}{${\overline{\operatorname{AUC}\textit{Skew}}}^*$} & CNN & 0.70$\pm$0.05 & 0.69$\pm$0.07 & \underline{0.95$\pm$0.05} & 0.68$\pm$0.06 & 0.00$\pm$0.00 & \textbf{0.99$\pm$0.01} \\
& BiLSTM & \underline{0.66$\pm$0.04} & 0.37$\pm$0.07 & 0.44$\pm$0.08 & 0.47$\pm$0.04 & 0.00$\pm$0.00 & \textbf{1.00$\pm$0.00} \\
& Trans & 0.48$\pm$0.15 & 0.58$\pm$0.19 & \underline{0.72$\pm$0.33} & 0.45$\pm$0.15 & 0.00$\pm$0.00 & \textbf{0.93$\pm$0.13} \\ 
\midrule
\multirow{3}{*}{${\operatorname{AUC}\textit{Kurt}}^*$} & CNN & 0.08$\pm$0.07 & 0.09$\pm$0.12 & 0.71$\pm$0.21 & 0.07$\pm$0.08 & \textbf{0.97$\pm$0.07} & \underline{0.76$\pm$0.16} \\
& BiLSTM & 0.09$\pm$0.06 & 0.04$\pm$0.03 & 0.03$\pm$0.02 & 0.00$\pm$0.01 & \textbf{0.99$\pm$0.02} & \underline{0.65$\pm$0.23} \\
& Trans & 0.10$\pm$0.13 & 0.30$\pm$0.26 & 0.56$\pm$0.42 & 0.19$\pm$0.11 & \underline{0.57$\pm$0.40} & \textbf{0.64$\pm$0.30} \\ 
\bottomrule
\end{tabular}
\end{sc}
\end{small}
\end{center}
\vskip -0.2in
\end{table*}

\setcounter{table}{0}
\section{Reproducibility of the results}\label{app:reproducibility}
In the supplementary material, we provide the code used to run all experiments, with the neural network architectures adopted and the interpretability framework used. We also provide the code to generate the synthetic dataset used in this work, along with the steps necessary to create it. We additionally provide instructions on how to download the other 20 public datasets used, and on how to format them to run with our framework.

\end{document}